\documentclass[11pt,a4paper]{article}
\pdfoutput=1
\usepackage[hyphens]{url}
\PassOptionsToPackage{breaklinks}{hyperref}
\usepackage[hyperref]{acl2019}
\usepackage{times}
\usepackage{latexsym}
\usepackage{multirow}
\usepackage{array}

\usepackage{graphicx}
\usepackage{booktabs}
\usepackage{numprint}
\usepackage{setspace}
\usepackage{multicol}
\usepackage{enumitem}

\usepackage[utf8]{inputenc}
\usepackage[T1]{fontenc}

\usepackage{xspace}
\usepackage{textcomp}

\usepackage{float}
\restylefloat{table}

\aclfinalcopy 

\newcommand\foursq{Foursquare}

\title{Machine Translation of Restaurant Reviews:\\ New Corpus for Domain Adaptation and Robustness}

\author{
	Alexandre B\'erard \qquad
	Ioan Calapodescu \qquad
	Marc Dymetman 
	\\[.4em]
	\hspace{.2em} {\bf Claude Roux} \qquad
	{\bf Jean-Luc Meunier} \qquad
	{\bf Vassilina Nikoulina} \\[.5em]
	Naver Labs Europe\\ \texttt{first.last@naverlabs.com}
	\\ \url{https://europe.naverlabs.com}
}

\date{}

\begin{document}
\maketitle

\begin{abstract}
We share a French-English parallel corpus of Foursquare restaurant reviews, and define a new task to encourage research on Neural Machine Translation robustness and domain adaptation, in a real-world scenario where better-quality MT would be greatly beneficial. We discuss the challenges of such user-generated content, and train good baseline models that build upon the latest techniques for MT robustness. We also perform an extensive evaluation (automatic and human) that shows significant improvements over existing online systems. Finally, we propose task-specific metrics based on sentiment analysis or translation accuracy of domain-specific polysemous words.
\end{abstract}

\section{Introduction}

Very detailed information about social venues such as restaurants is available from user-generated reviews in applications like Google Maps, TripAdvisor or Foursquare.\footnote{\url{https://foursquare.com/}}
Most of these reviews are written in the local language and are not directly exploitable by foreign visitors: an analysis of the \foursq{} database shows that, in Paris, only 49\% of the restaurants have at least one review in English. It can be much worse for other cities and languages (e.g., only 1\% of Seoul restaurants for a French-only speaker).

Machine Translation of such user-generated content can improve the situation and make the data available for direct display or for downstream NLP tasks (e.g., cross-lingual information retrieval, sentiment analysis, spam or fake review detection), provided its quality is sufficient.

We asked professionals to translate 11.5k French \foursq{} reviews (18k sentences) to English. We believe that this resource\footnote{\url{https://europe.naverlabs.com/research/natural-language-processing/machine-translation-of-restaurant-reviews/}} will be valuable to the community for training and evaluating MT systems addressing challenges posed by user-generated content, which we discuss in detail in this paper.

We conduct extensive experiments and combine techniques that seek to solve these challenges (e.g., factored case, noise generation, domain adaptation with tags) on top of a strong Transformer baseline.
In addition to BLEU evaluation and human evaluation, we use targeted metrics that measure how well polysemous words are translated, or how well sentiments expressed in the original review can still be recovered from its translation.


\section{Related work}
\label{section:related_work}

Translating restaurant reviews written by casual customers presents several difficulties for NMT, in particular robustness to non-standard language and adaptation to a specific style or domain (see Section~\ref{sec:challenges} for details).

Concerning robustness to noisy user generated content, \citet{michel_mtnt:_2018} stress differences with traditional domain adaptation problems, and propose a typology of errors, many of which we also detected in the \foursq{} data.
They also released a dataset (MTNT), whose sources were selected from a social media (Reddit) on the basis of being especially noisy (see Appendix for a comparison with \foursq{}).
These sources were then translated by humans to produce a parallel corpus that can be used to engineer more robust NMT systems and to evaluate them. This corpus was the basis of the WMT 2019 Robustness Task \cite{WMT_robustness_2019}, in which \citet{berard:2019:WMT} ranked first. We use the same set of robustness and domain adaptation techniques, which we study more in depth and apply to our review translation task.

\citet{Sperber2017}, \citet{belinkov_synthetic_2018} and \citet{karpukhin_2019} propose to improve robustness by training models on data-augmented corpora, containing noisy sources obtained by random word or character deletions, insertions, substitutions or swaps.
Recently, \citet{vaibhav_2019} proposed to use a similar technique along with noise generation through replacement of a clean source by one obtained by back-translation.

We employ several well-known domain adaptation techniques: back-translation of large monolingual corpora close to the domain \cite{sennrich_improving_2016,edunov_understanding_2018}, fine-tuning with in-domain parallel data \cite{luong_stanford_2015,freitag_fast_2016,servan_2016}, domain tags for knowledge transfer between domains \cite{kobus_domain_2017,berard:2019:WMT}.

Addressing the technical issues of robustness and adaptation of an NMT system is decisive for real-world deployment, but evaluation is also critical.
This aspect is stressed by \citet{levin_toward_2017} (NMT of curated hotel descriptions), who point out that automatic metrics like BLEU tend to neglect semantic differences that have a small textual footprint, but may be seriously misleading in practice, for instance by interpreting \textit{available parking} as if it meant \textit{free parking}.
To mitigate this, we conduct additional evaluations of our models: human evaluation, translation accuracy of polysemous words, and indirect evaluation with sentiment analysis.

\section{Task description}
\label{section:task_description}

We present a new task of restaurant review translation, which combines domain adaptation and robustness challenges.

\subsection{Corpus description}

We sampled 11.5k French reviews from \foursq{}, mostly in the \emph{food} category,\footnote{\url{https://developer.foursquare.com/docs/resources/categories}} split them into 18k sentences, and grouped them into train, valid and test sets (see Table~\ref{table4SQ}). The French reviews contain on average 1.5 sentences and 17.9 words. Then, we hired eight professional translators to translate them to English.
Two of them created the training set by post-editing (PE) the outputs of baseline NMT systems.\footnote{ConvS2S or Transformer Big trained on the ``UGC'' corpus described in Section~\ref{section:experiments}, without domain adaptation or robustness tricks.} The other six translated the valid and test sets from scratch. They were asked to translate (or post-edit) the reviews sentence-by-sentence (to avoid any alignment problem), but they could see the full context.
We manually filtered the test set to remove translations that were not satisfactory. The full reviews and additional metadata (e.g., location and type of the restaurant) are also available as part of this resource, to encourage research on contextual machine translation.

\foursq{}-HT was translated from scratch by the same translators who post-edited \foursq{}-PE. While we did not use it in this work, it can be used as extra training or development data. We also release a human translation of the French-language test set (668 sentences) of the Aspect-Based Sentiment Analysis task at SemEval 2016 \cite{8131987}.

\begin{table}
	\centering
	\begin{tabular}{l|ccc}
		Split & Sentences & Reviews & Words (FR) \\
		\hline
		PE (train) & \numprint{12080} & \numprint{8004} & \numprint{141958} \\
		HT & \numprint{2784} & \numprint{1625} & \numprint{29075} \\		
		valid & \numprint{1243} & 765 & \numprint{13976} \\
		test & \numprint{1838} & \numprint{1157} & \numprint{21525} \\
	\end{tabular}
	\vspace{-.1cm}
	\caption{\foursq{} splits. \foursq{}-PE is the training set. \foursq{}-HT is not used in this work.}
	\label{table4SQ}	
\end{table}

\subsection{Challenges}
\label{sec:challenges}

\vspace{-.2cm}

\begin{center}
	\begin{small}
		\begin{tabular}{rp{4.5cm}l}
			\multirow{3}{*}{(1)} & é qd g vu sa ...   & (source) \\
			& and when I saw that ... & (reference) \\
			& \textcolor{red}{é qd g seen his ...} & (online MT) \\
			\cmidrule{2-2}
			\multirow{3}{*}{(2)} & c'est trooop bon ! & \\
			& it's toooo good! & \\
			& \textcolor{red}{it's good trooop!} & \\
			\cmidrule{2-2}
			\multirow{3}{*}{(3)} & le cadre est nul & \\
			& the setting is lousy & \\
			& \textcolor{red}{the frame is null} & \\
			\cmidrule{2-2}
			\cmidrule{2-2}
			\multirow{3}{*}{(4)} & le garçon a pété un cable & \\
			& the waiter went crazy & \\
			& \textcolor{red}{the boy farted a cable} & \\
			\cmidrule{2-2}
			\multirow{3}{*}{(5)} & pizza nickel, tres bonnes pattes & \\
			& great pizza, very good pasta & \\
			& \textcolor{red}{nickel pizza, very good legs} & \\
		\end{tabular}
	\end{small}
\end{center}

Translating restaurant reviews presents two main difficulties compared to common tasks in MT.
First, the reviews are written in a casual style, close to spoken language.
Some liberty is taken w.r.t. spelling, grammar, and punctuation. Slang is also very frequent.
MT should be robust to these variations.
Second, they generally are reactions, by clients of a restaurant, about its food quality, service or atmosphere, with specific words relating to these aspects or sentiments.
These require some degree of domain adaptation.
The table above illustrates these issues, with outputs from an online MT system. Examples of full reviews from \foursq{}-PE along with metadata are shown in Appendix.

Examples 1 and 2 fall into the robustness category: 1 is an extreme form of SMS-like, quasi-phonetic, language (\textit{et quand j'ai vu ça}); 2 is a literal transcription of a long-vowel phonetic stress (\textit{trop} $\rightarrow$ \textit{trooop}).
Example 3 falls into the domain category: in a restaurant context, \textit{cadre} typically refers to the \textit{setting}.
Examples 4 and 5 involve both robustness and domain adaptation: \textit{pété un cable} is a non-compositional slang expression and \textit{garçon} is not a \textit{boy} in this domain; \textit{nickel} is slang for \textit{great}, \textit{très} is missing an accent, and \textit{pâtes} is misspelled as \textit{pattes}, which is another French word.

Regarding robustness, we found many of the same errors listed by \citet{michel_mtnt:_2018} as noise in social media text: SMS language (\textit{é qd g vu sa}), typos and phonetic spelling (\textit{pattes}), repeated letters (\textit{trooop, merciiii}), slang (\textit{nickel, bof, mdr}), missing or wrong accents (\textit{tres}), emoticons (`:-)') and emojis,
missing punctuation, wrong or non-standard capitalization (lowercase proper names, capitalized words for emphasis).
Regarding domain aspects, there are polysemous words with typical specific meaning \textit{carte} $\rightarrow$ \textit{map}, \underline{\textit{menu}}; \textit{cadre} $\rightarrow$ \textit{frame}, \textit{executive}, \underline{\textit{setting}}), idiomatic expressions (\textit{à tomber par terre} $\rightarrow$ \textit{to die for}), and venue-related named entities (\textit{La Boîte à Sardines}).

\section{Robustness to noise}
\label{section:robustness}

We propose solutions for dealing with non-standard case, emoticons, emojis and other issues.

\subsection{Rare character placeholder}
\label{section:rare_chars}

We segment our training data into subwords with BPE \citep{sennrich_neural_2016}, implemented in SentencePiece \citep{kudo_sentencepiece:_2018}.
BPE can deal with rare or unseen words by splitting them into more frequent subwords, but cannot deal with unseen characters.\footnote{Unless actually doing BPE at the \emph{byte} level, as suggested by \citet{radford_2019}.} While this is not a problem in most tasks, \foursq{} contains many emojis, and sometimes symbols in other scripts (e.g., Arabic).
Unicode now defines around 3k emojis, most of which are likely to be out-of-vocabulary.

We replace rare characters on both sides of the training corpus by a placeholder (\texttt{<x>}). A model trained on this data is typically able to copy the placeholder at the correct position.
Then, at inference time, we replace the output tokens \texttt{<x>} by the rare source-side characters, in the same order.
This approach is similar to that of \citet{jean_using_2015}, who used the attention mechanism to replace \texttt{UNK} symbols with the aligned word in the source. \citet{berard:2019:WMT} used the same technique to deal with emojis in the WMT robustness task.

\subsection{Capital letters}
\label{section:case_handling}

As shown in Table~\ref{table:capital_letters_break_nmt}, capital letters are another source of confusion.
\textit{HONTE} and \textit{honte} are considered as two different words.
The former is out-of-vocabulary and is split very aggressively by BPE.
This causes the MT model to hallucinate.

\begin{table}
	\hspace{-.5cm}
	\small
	\begin{tabular}{lll}
		& Uppercase & Lowercase \\
		Input & \tt UNE \textbf{HONTE} ! & \tt une \textbf{honte} !\\
		Pre-proc & \tt UN E \_H ON TE \_! & \tt une \_honte \_!\\
		MT output & \tt A \_H ON E Y ! & \tt A \_dis gra ce ! \\
		Post-proc & \tt A \textbf{HONEY}! & \tt A \textbf{disgrace}! \\
	\end{tabular}
	\vspace{-.1cm}
	\caption{Capital letters break NMT.
	BPE segmentation and translation of capitalized or lowercase input.}
	\label{table:capital_letters_break_nmt}
\end{table}

\paragraph{Lowercasing}
A solution is to lowercase the input, both at training and at test time.
However, when doing so, some information may be lost (e.g., named entities, acronyms, emphasis) which may result in lower translation quality.

\paragraph{Factored translation}
\citet{levin_toward_2017} do factored machine translation \cite{sennrich_linguistic_2016,garcia-martinez_factored_2016} where a word and its case are split in two different features.
For instance, \textit{HONTE} becomes \textit{honte + upper}.

We implement this with two embedding matrices, one for words and one for case, and represent a token as the sum of the embeddings of its factors.
For the target side, we follow \citet{garcia-martinez_factored_2016} and have two softmax operations.
We first predict the word in its lowercase form and then predict its case.\footnote{Like the ``dependency model'' of \citet{garcia-martinez_factored_2016}, we use the current state of the decoder and the embedding of the output word to predict its case.} The embeddings of the case and word are then summed and used as input for the next decoder step.

\paragraph{Inline casing}
\citet{berard:2019:WMT} propose another approach, \emph{inline casing}, which does not require any change in the model.
We insert the case as a regular token into the sequence right after the word.
Special tokens \texttt{<U>}, \texttt{<L>} and \texttt{<T>} (upper, lower and title) are used for this purpose and appended to the vocabulary.
Contrary to the previous solution, there is only one embedding matrix and one softmax.

In practice, words are assumed to be lowercase by default and the \texttt{<L>} tokens are dropped to keep the factored sequences as short as possible.
\textit{``Best fries EVER"} becomes \textit{``best <T> \_f ries \_ever <U>"}. Like \citet{berard:2019:WMT}, we force SentencePiece to split mixed-case words like \textit{MacDonalds} into single-case subwords (\textit{Mac} and \textit{Donalds}).

\paragraph{Synthetic case noise}
Another solution that we experiment with (see Section~\ref{section:experiments}) is to inject noise on the source side of the training data by changing random source words to upper (5\% chance), title (10\%) or lower case (20\%).

\subsection{Natural noise}
\label{section:natural_noise}
One way to make an NMT system more robust is to train it with some of the most common errors that can be found in the in-domain data.
Like \citet{berard:2019:WMT}, we detect the errors that occur naturally in the in-domain data and then apply them to our training corpus, while respecting their natural distribution.
We call this ``natural noise generation'' in opposition to what is done in \cite{Sperber2017,belinkov_synthetic_2018,vaibhav_2019} or in Section~\ref{section:case_handling}, where the noise is more synthetic.

\paragraph{Detecting errors}
We compile a general-purpose French lexicon as a transducer,\footnote{In Tamgu: \url{https://github.com/naver/tamgu}} implemented to be traversed with extended edit distance flags, similar to \citet{mihov_fast_2004}.
Whenever a word is not found in the lexicon (which means that it is a potential spelling mistake), we look for a French word in the lexicon within a maximum edit distance of 2, with the following set of edit operations:

\begin{center}
	\begin{small}
		\begin{tabular}{rp{5.5cm}}
			(1) & deletion (e.g., \textit{apelle} instead of \textit{appelle}) \\
			\cmidrule{2-2}
			(2) & insertion (e.g., \textit{appercevoir} instead of \textit{apercevoir}) \\
			\cmidrule{2-2}
			(3) & constrained substitution on diacritics (e.g., \textit{mangè} instead of \textit{mangé}) \\
			\cmidrule{2-2}
			(4) & swap counted as one operation: (e.g., \textit{mnager} instead of \textit{manger}) \\
			\cmidrule{2-2}
			(5) & substitution  (e.g., \textit{menger} instead of \textit{manger}) \\
			\cmidrule{2-2}
			(6) & repetitions (e.g., \textit{Merciiiii} with a threshold of max 10 repetitions) \\
		\end{tabular}
	\end{small}
\end{center}

We apply the transducer to the French monolingual Foursquare data (close to 1M sentences) to detect and count noisy variants of known French words.
This step produces a dictionary mapping the correct spelling to the list of observed errors and their respective frequencies.

In addition to automatically extracted spelling errors, we extract a set of common abbreviations from \cite{seddah_2012} and we manually identify a list of common errors in French:

\begin{center}
	\begin{small}
		\begin{tabular}{rp{5.5cm}}
			(7) & Wrong verb endings (e.g., \textit{il a manger} instead of \textit{il a mangé}) \\
			\cmidrule{2-2}
			(8) & Wrong spacing around punctuation symbols (e.g., \textit{Les.plats ...} instead of \textit{Les plats...}) \\
			\cmidrule{2-2}
			(9) & Upper case/mixed case words (e.g., \textit{manQue de place} instead of \textit{manque de place}) \\
			\cmidrule{2-2}
			(10) & SMS language (e.g., \textit{bcp} instead of \textit{beaucoup}) \\
			\cmidrule{2-2}
			(11) & Phonetic spelling (e.g., \textit{sa} instead of \textit{ça}) \\
		\end{tabular}
	\end{small}
\end{center}

\paragraph{Generating errors}

With this dictionary, describing the real error distribution in \foursq{} text, we take our large out-of-domain training corpus, and randomly replace source-side words with one of their variants (rules 1 to 6), while respecting the frequency of this variant in the real data.
We also manually define regular expressions to randomly apply rules 7 to 11 (e.g., \texttt{"er "}$\to$\texttt{"é "}).

We obtain a noisy parallel corpus (which we use instead of the ``clean'' training data), where about 30\% of all source sentences have been modified, as shown below:

\vspace{-.3cm}

\begin{center}
	\begin{small}
		\begin{tabular}{rp{5.5cm}}
			Error type & Examples of sentences with injected noise\\
			\cmidrule{2-2}
			(1) (6) (9) & L'Union \textbf{eU}ropée\textbf{ne} esp\textbf{ere} que la réunion de suiv\textbf{iii} entre le Président [...]\\
			\cmidrule{2-2}
			(2) (3) (10) & Le Comité no\textbf{tte} avec \textbf{bcp} d'int\textbf{eret} \textbf{k} les projets d'articles [...] \\
			\cmidrule{2-2}
			(4) (7) (8) & Réun\textbf{oin} \textbf{sur.la} comptabilit\textbf{er} nationale [...]\\
		\end{tabular}
	\end{small}
\end{center}

\section{Domain Adaptation}
\label{section:domain_adaptation}

To adapt our models to the restaurant review domain we apply the following types of techniques: back-translation of in-domain English data, fine-tuning with small amounts of in-domain parallel data, and domain tags.

\subsection{Back-translation}
\label{section:back_translation}

Back-translation (BT) is a popular technique for domain adaptation when large amounts of in-domain monolingual data are available \citep{sennrich_improving_2016,edunov_understanding_2018}.
While our in-domain parallel corpus is small (12k pairs), Foursquare contains millions of English-language reviews.
Thus, we train an NMT model\footnote{Like the ``UGC'' model with rare character handling and inline case described in Section~\ref{section:model_and_settings}.} in the reverse direction (EN$\to$FR) and translate all the \foursq{} English reviews to French.\footnote{This represents $\approx$15M sentences.
This corpus is not available publicly, but the Yelp dataset (\url{https://www.yelp.com/dataset}) could be used instead.}
This gives a large synthetic parallel corpus.

This \textit{in-domain} data is concatenated to the out-of-domain parallel data and used for training.

\citet{edunov_understanding_2018} show that doing back-translation with sampling instead of beam search brings large improvements due to increased diversity.
Following this work, we test several settings:

\begin{center}
	\begin{small}
		\begin{tabular}{rp{5.5cm}}
			Name & Description\\
			\cmidrule{2-2}
			BT-B & Back-translation with beam search. \\
			\cmidrule{2-2}
			BT-S & Back-translation with sampling.  \\
			\cmidrule{2-2}
			BT-S $\times$ 3 & Three different FR samplings for each EN sentence. This brings the size of the back-translated \foursq{} closer to the out-of-domain corpus. \\
			\cmidrule{2-2}
			BT & No oversampling, but we sample a new version of the corpus for each training epoch. \\
		\end{tabular}
	\end{small}
\end{center}

We use a temperature\footnote{with $p(w_i) = \frac{exp(z_i/T)}{\sum_{k=1}^{|V|}{exp(z_k/T)}}$} of $T=\frac{1}{0.9}$ to avoid the extremely noisy output obtained with $T=1$ and strike a balance between quality and diversity.

\subsection{Fine-tuning}

When small amounts of in-domain parallel data are available, fine-tuning (FT) is often the preferred solution for domain adaptation \citep{luong_stanford_2015,freitag_fast_2016}.
It consists in training a model on out-of-domain data, and then continuing its training for a few epochs on the in-domain data only.

\subsection{Corpus tags}

\citet{kobus_domain_2017} propose a technique for multi-domain NMT, which consists in inserting a token in each source sequence specifying its domain.
The system can learn the particularities of multiple domains (e.g., polysemous words that have a different meaning depending on the domain), which we can control at test time by manually setting the tag.
\citet{sennrich_controlling_2016} also use tags to control politeness in the model's output.

As our corpus (see Section~\ref{section:training_data}) is not clearly divided into  domains, we apply the same technique as \citet{kobus_domain_2017} but use \emph{corpus} tags (each sub-corpus has its own tag: \texttt{TED}, \texttt{Paracrawl}, etc.) which we add to each source sequence.
Like in \cite{berard:2019:WMT}, the \foursq{} post-edited and back-translated data also get their own tags (\texttt{PE} and \texttt{BT}).
Figure \ref{fig:tag_ex} gives an example where using the \texttt{PE} corpus tag at test time helps the model pick a more adequate translation.
\begin{figure}
	\hspace{-.4cm}
	\small
	\begin{tabular}{rp{5.5cm}}
		Corpus tag & SRC: La \textbf{carte} est trop petite.\\
		\cmidrule{2-2}
		\texttt{TED} & The map is too small. \\
		\cmidrule{2-2}
		\texttt{Multi-UN} & The card is too small. \\
		\cmidrule{2-2}
		\texttt{PE} & The \textbf{menu} is too small. \\
	\end{tabular}
	\vspace{-.2cm}
	\caption{Example of ambiguous source sentence, where using corpus tags helps the model pick a more adequate translation.}
	\label{fig:tag_ex}
\end{figure}

\section{Experiments}
\label{section:experiments}

\subsection{Training data}
\label{section:training_data}

After some initial work with the WMT 2014 data, we built a new training corpus named UGC (User Generated Content), closer to our domain, by combining: Multi UN, OpenSubtitles, Wikipedia, Books, Tatoeba, TED talks, ParaCrawl\footnote{Corpora available at \url{http://opus.nlpl.eu/}} and Gourmet\footnote{3k translations of dishes and other food terminology \url{http://www.gourmetpedia.eu/}} (See Table~\ref{table:ugc_size}).
UGC does not include Common Crawl (which contains many misaligned sentences and caused hallucinations), but it includes OpenSubtitles \cite{lison_opensubtitles2016:_2016} (spoken-language, possibly closer to \foursq{}).
We observed an improvement of more than 1 BLEU on newstest2014 when switching to UGC, and almost 6 BLEU on \foursq{}-valid.

\begin{table}
	\centering
	\begin{tabular}{c|c|c|c}
		Corpus & Lines & Words (FR) & Words (EN) \\
		\hline
		WMT & 29.47M & 1 003M & 883.5M \\
		UGC & 51.39M & 1 125M & 1 041M \\
	\end{tabular}
	\vspace{-.2cm}
	\caption{Size of the WMT and UGC training corpora (after filtering).}
	\label{table:ugc_size}
\end{table}

\subsection{Pre-processing}
We use \texttt{langid.py} \citep{lui_2012} to filter sentence pairs from UGC.
We also remove duplicate sentence pairs, and lines longer than $175$ words or with a length ratio greater than $1.5$ (see Table~\ref{table:ugc_size}).
Then we apply SentencePiece and our rare character handling strategy (Section~\ref{section:rare_chars}).
We use a joined BPE model of size 32k, trained on the concatenation of both sides of the corpus, and set SentencePiece's vocabulary threshold to $100$.
Finally, unless stated otherwise, we always use the \emph{inline casing} approach (see Section~\ref{section:case_handling}).

\subsection{Model and settings}
\label{section:model_and_settings}
For all experiments, we use the Transformer Big \cite{vaswani_attention_2017} as implemented in Fairseq, with the hyperparameters of \citet{ott_scaling_2018}.
Training is done on 8 GPUs, with accumulated gradients over 10 batches \citep{ott_scaling_2018}, and a max batch size of $3500$ tokens per GPU.
We train for $20$ epochs, while saving a checkpoint every $2500$ updates ($\approx\frac{2}{5}$ epoch on UGC) and average the 5 best checkpoints according to their perplexity on a validation set (a held-out subset of UGC).

For fine-tuning, we use a fixed learning rate, and a total batch size of 3500 tokens (training on a single GPU without delayed updates).
To avoid overfitting on \foursq{}-PE, we do early stopping according to perplexity on \foursq{}-valid.\footnote{The best perplexity was achieved after 1 to 3 epochs.}
For each fine-tuned model we test all 16 combinations of dropout in $\{0.1,0.2,0.3,0.4\}$ and learning rate in $\{1, 2, 5, 10\}\times10^{-5}$.
We keep the model with the best perplexity on \foursq{}-valid.\footnote{The best dropout rate was always $0.1$, and the best learning rate was either $2\times10^{-5}$ or $5\times10^{-5}$.}

\subsection{Evaluation methodology}
During our work, we used BLEU \citep{papineni_bleu:_2002} on newstest[2012, 2013] to ensure that our models stayed good on a more general domain, and on \foursq{}-valid to measure performance on the \foursq{} domain.

For sake of brevity, we only give the final BLEU scores on newstest2014 and \foursq{}-test. Scores on \foursq{}-valid, and MTNT-test (for comparison with \citealp{michel_mtnt:_2018,berard:2019:WMT}) are given in Appendix. We evaluate ``detokenized'' MT outputs\footnote{Outputs of our models are provided with the \foursq{} corpus.} against raw references using SacreBLEU \citep{post_call_2018}.\footnote{SacreBLEU signature: BLEU+case.mixed+numrefs.1\\+smooth.exp+tok.13a+version.1.2.10}

In addition to BLEU, we do an indirect evaluation on an Aspect-Based Sentiment Analysis (ABSA) task, a human evaluation, and a task-related evaluation based on polysemous words.

\subsection{BLEU evaluation}

\paragraph{Capital letters}
\begin{table}[t]
	\hspace{-.3cm}
	\begin{tabular}{l|c|c|c|c}
		\multirow{2}{*}{Model} & \multirow{2}{*}{BLEU} & \multicolumn{3}{c}{Case insensitive BLEU} \\
		& & Upper & Lower & Title \\
		\hline
		Cased                & 31.75          & 16.01          & 32.39          & 26.66 \\ 
		LC to cased          & 30.70          & \textbf{33.03} & 33.03          & 33.03 \\ 
		Factored case        & 31.59          & 32.25          & 32.96          & 29.83 \\ 
		Inline case          & 31.46          & 31.08          & 32.57          & 29.55 \\ 
		Noised case          & \textbf{31.83} & 32.61          & \textbf{33.69} & \textbf{33.60} \\ 
	\end{tabular}
	\vspace{-.2cm}
	\caption{Robustness to capital letters (see Section~\ref{section:case_handling}).
	\foursq{}-test's source side has been set to upper, lower or title case.
	The first column is case sensitive BLEU on \foursq{}-test.
	``LC to cased'' always gets the same scores because it is invariant to source case.}
	\label{table:BLEU_case}
\end{table}

Table~\ref{table:BLEU_case} compares the case handling techniques presented in Section~\ref{section:case_handling}.
To better evaluate the robustness of our models to changes of case, we built 3 synthetic test sets from \foursq{}-test, with the same target, but all source words in upper, lower or title case.

Inline and factored case perform equally well, significantly better than the default (cased) model, especially on all-uppercase inputs.
Lowercasing the source is a good option, but gives a slightly lower score on regular \foursq{}-test.\footnote{The ``LC to cased'' and ``Noised case'' models are not able to preserve capital letters for emphasis (as in Table~\ref{table:capital_letters_break_nmt}), and the ``Cased'' model often breaks on such examples.} Finally, synthetic case noise added to the source gives surprisingly good results.
It could also be combined with factored or inline case.

\paragraph{Natural noise}
Table~\ref{table:BLEU_nat_noise} compares the baseline ``inline case'' model with the same model augmented with natural noise (Section~\ref{section:natural_noise}).
Performance is the same on \foursq{}-test, but significantly better on newstest2014 artificially augmented with \foursq{}-like noise.

\begin{table}[t]
	\hspace{-.3cm}
	\begin{tabular}{l|c|c|c}
		Model & news & noised news & test \\
		\hline
		UGC (Inline case) & \textbf{40.68} & 35.59 & 31.46 \\ 
		+ natural noise & 40.43 & \textbf{40.35} & \textbf{31.66} \\ 
	\end{tabular}
	\vspace{-.2cm}
	\caption{Baseline model with or without natural noise (see Section~\ref{section:natural_noise}).
	\emph{Noised news} is the same type of noise, artificially applied to newstest2014.}
	\label{table:BLEU_nat_noise}
\end{table}

\paragraph{Domain adaptation}

Table~\ref{table:BLEU_back-translation} shows the results of the back-translation (BT) techniques.
Surprisingly, BT with beam search (BT-B) deteriorates BLEU scores on \foursq{}-test, while BT with sampling gives a consistent improvement.
BLEU scores on newstest2014 are not significantly impacted, suggesting that BT can be used for domain adaptation without hurting quality on other domains.

\begin{table}[t]
	\centering
	\begin{tabular}{l|c|c}
		Model & news & test \\
		\hline
		UGC (Inline case) & 40.68 & 31.46 \\  
		UGC $\oplus$ BT-B & 40.56 & 30.14 \\   
		UGC $\oplus$ BT-S & 40.64 & 32.59 \\   
		UGC $\oplus$ BT & \textbf{40.84} & 32.68 \\  
		UGC $\oplus$ BT-S $\times$ 3 & 40.63 & \textbf{32.80} \\  
	\end{tabular}
	\vspace{-.2cm}
	\caption{Comparison of different back-translation schemes (see Section~\ref{section:back_translation}).
	$\oplus$ denotes the concatenation of several training corpora.}
	\label{table:BLEU_back-translation}
\end{table}

\begin{table}[t]
	\centering
	\begin{tabular}{l|c|c|c}		
		Model & Tag & news & test \\
		\hline
		UGC (Inline case) & -- & 40.68 & 31.46 \\ 
		UGC $\oplus$ PE & -- & 40.80 & 31.98 \\ 
		UGC + FT & -- & 39.78 & \textbf{34.97} \\ 
		\hline
		\multirow{2}{*}{UGC $\oplus$ PE + tags} & -- & 40.71 & 32.15 \\ 
		& \texttt{PE} & 38.97 & 34.30 \\	
		\hline
		\multirow{2}{*}{UGC $\oplus$ BT + tags} & -- & 40.67 & 33.44 \\ 
		& \texttt{BT} & 39.02 & 32.87 \\
	\end{tabular}
	\vspace{-.2cm}
	\caption{Domain adaptation with \foursq{}-PE fine-tuning (FT) or corpus tags.
	The ``tag'' column represents the corpus tag used at test time (if any).}
	\label{table:BLEU_domain_adaptation}
\end{table}

Table~\ref{table:BLEU_domain_adaptation} compares the domain adaptation techniques presented in Section~\ref{section:domain_adaptation}.
We observe that:

\vspace{-.1cm}

\begin{enumerate}[leftmargin=.5cm]
\itemsep0em
\item Concatenating the small \foursq{}-PE corpus to the 50M general domain corpus does not help much, unless using corpus tags.
\item \emph{\foursq{}-PE + tags} is not as good as fine-tuning with \foursq{}-PE.
However, fine-tuned models get slightly worse results on news.
\item Back-translation combined with tags gives a large boost.\footnote{\citet{Caswell2019,berard:2019:WMT} observed the same thing.} The \texttt{BT} tag should not be used at test time, as it degrades results.
\item Using no tag at test time works fine, even though all training sentences had tags.\footnote{We tried keeping a small percentage of UGC with no tag, or with an \texttt{ANY} tag, but this made no difference.}
\end{enumerate}

\begin{table}[t]
	\centering
	\begin{tabular}{l|c|c}
		Model & news & test \\
		\hline
		WMT & 39.37 & 26.23 \\ 
		UGC (Inline case) & 40.68 & 31.46 \\ 
		Google Translate (Feb 2019) & 36.31 & 29.63 \\
		DeepL (Feb 2019) & ? & 32.82 \\
		\hline
		UGC $\oplus$ BT + FT & 39.55 & 35.93 \\ 
		UGC $\oplus$ BT $\oplus$ PE + tags & \textbf{40.99} & 35.60 \\ 
		Nat noise $\oplus$ BT + FT & 39.91 & \textbf{36.25} \\ 
		Nat noise $\oplus$ BT $\oplus$ PE + tags & 40.72 & 35.54 \\ 
	\end{tabular}
	\vspace{-.2cm}
	\caption{Combination of several robustness or domain adaptation techniques.
	At test time, we don't use any tag on news, and use the \texttt{PE} tag on \foursq{}-test (when applicable).
	BT: back-translation.
	PE: \foursq{}-PE.
	FT: fine-tuning with Foursquare-PE.
	$\oplus$: concatenation.}
	\label{table:BLEU_combination}
\end{table}

As shown in Table~\ref{table:BLEU_combination}, these techniques can be combined to achieve the best results.
The natural noise does not have a significant effect on BLEU scores.
Back-translation combined with fine-tuning gives the best performance on \foursq{} (+4.5 BLEU vs UGC).
However, using tags instead of fine-tuning strikes a better balance between general domain and in-domain performance.

\subsection{Targeted evaluation}

In this section  we propose two metrics that target specific aspects of translation adequacy: translation accuracy of domain-specific polysemous words and Aspect-Based Sentiment Analysis performance on MT outputs.

\paragraph{Translation of polysemous words}

We propose to count polysemous words specific to our domain, similarly to \citet{lala_2018}, to measure the degree of domain adaptation.
TER between the translation hypotheses and the post-edited references in \foursq{}-PE reveals the most common substitutions (e.g., ``card'' is often replaced with ``menu'', suggesting that ``card'' is a common mistranslation of the polysemous word ``carte'').
We filter this list manually to only keep words that are polysemous and that have a high frequency in the test set.
Table~\ref{table:polysemous_words} gives the 3 most frequent ones.\footnote{Rarer ones are: \emph{adresse} (place, address), \emph{café} (coffee, café), \emph{entrée} (starter, entrance), \emph{formule} (menu, formula), \emph{long} (slow, long), \emph{moyen} (average, medium), \emph{correct} (decent, right), \emph{brasserie} (brasserie, brewery) and \emph{coin} (local, corner).}

Table~\ref{table:polysemous_accuracy} shows the accuracy of our models when translating these words.
We see that the domain-adapted model is better at translating domain-specific polysemous words.

\begin{table}
	\centering
	\begin{tabular}{l|l}
		French word & Meanings \\
		\hline
		Cadre & \underline{setting}, frame, executive \\
		Cuisine & \underline{food}, kitchen \\
		Carte & \underline{menu}, \underline{card}, map
	\end{tabular}
	\vspace{-.2cm}
	\caption{French polysemous words found in \foursq{}, and translation candidates in English.
	The most frequent meanings in \foursq{} are underlined.}
	\label{table:polysemous_words}
\end{table}

\begin{table}[t]
	\hspace{-.4cm}
	\begin{tabular}{l|ccc|c}
		Model & cadre & cuisine & carte & Total \\
		\hline
		Total (source) & 23 & 32 & 29 & 100\% \\
		\hline
		WMT & 13 & 17 & 14 & 52\% \\ 
		UGC (Inline case) & 22 & 27 & 18 & 80\% \\ 
		\hline
		UGC $\oplus$ PE + tags & \textbf{23} & 31 & \textbf{29} & 99\% \\ 
	\end{tabular}
	\vspace{-.15cm}
	\caption{Number of correct translations for difficult polysemous words in \foursq{}-test by different models.
	The first row is the number of source sentences that contain this word.
	Other domain-adapted models (e.g., ``UGC + FT'' or ``UGC $\oplus$ BT'') also get $\approx$ 99\% accuracy.}
	\label{table:polysemous_accuracy}
\end{table}

\paragraph{Indirect evaluation with sentiment analysis}

We also measure adequacy by how well the translation preserves the polarity of the sentence regarding various aspects.
To evaluate this, we perform an indirect evaluation on the SemEval 2016 Aspect-Based Sentiment Analysis  (ABSA) task \cite{8131987}.
We use our internal ABSA systems trained on English or French SemEval 2016 data.
The evaluation is done on the SemEval 2016 French test set: either the original version (ABSA French), or its translation (ABSA English).
As shown in Table~\ref{table:ABSA}, translations obtained with domain-adapted models lead to significantly better scores on ABSA than the generic models.

\begin{table}
	\centering	
	\begin{tabular}{l|c|c}
		ABSA Model & Aspect  & Polarity \\
		\hline
		\emph{ABSA French} & \textbf{64.7} & \textbf{83.2}  \\ 
		\emph{ABSA English} & 59.5 & 72.1 \\
		\hline
		\multicolumn{3}{c}{\emph{ABSA English} on MT outputs}\\\hline
		WMT & 54.5 &  66.1  \\
		UGC (Inline case) & 58.1 & 70.7 \\ 
		UGC $\oplus$ BT $\oplus$ PE + tags & 60.2  & 72.0 \\ 
		Nat noise $\oplus$ BT $\oplus$ PE + tags & 60.8 & 73.3  \\ 
	\end{tabular}
	\vspace{-.15cm}
	\caption{Indirect evaluation with Aspect-Based Sentiment Analysis (accuracy in \%). \emph{ABSA French}: ABSA model trained on French data and applied to the SemEval 2016 French test set; \emph{ABSA English}: trained on English data and applied to human translations of the test set; \emph{ABSA English} on MT outputs: applied to MT outputs instead of human translations.}
	\label{table:ABSA}
\end{table}

\newcommand\baseline{Baseline}
\newcommand\google{GT}
\newcommand\noise{Tags + noise}
\newcommand\tags{Tags}

\subsection{Human Evaluation}
\label{section:human_eval}

We conduct a human evaluation to confirm the observations with BLEU and to overcome some of the limitations of this metric.

We select 4 MT models for evaluation (see Table~\ref{table_NLE_results}) and show their 4 outputs at once, sentence-by-sentence, to human judges, who are asked to rank them given the French source sentence in context (with the full review). 
For each pair of models, we count the number of wins, ties and losses, and apply the Wilcoxon signed-rank test.

We took the first 300 test sentences to create 6 tasks of 50 sentences each.
Then we asked bilingual colleagues to rank the output of 4 models by their translation quality.
They were asked to do one or more of these tasks.
The judge did not know about the list of models, nor the model that produced any given translation.
We got 12 answers.
The inter-judge Kappa coefficient ranged from 0.29 to 0.63, with an average of 0.47, which is a good value given the difficulty of the task. Table~\ref{table_NLE_results} gives the results of the evaluation, which confirm our observations with BLEU.

We also did a larger-scale monolingual evaluation using Amazon Mechanical Turk (see Appendix), which lead to similar conclusions.

\begin{table}[t]
	\centering	
	\begin{tabular}{l|c|c|c}
		Pairs & Win & Tie & Loss \\
		\hline
		\tags{} $\approx$ \noise{} & 82  & 453 & 63 \\
		\tags{} $\gg$ \baseline{}  & 187 & 337 & 74 \\
		\tags{} $\gg$ \google{}     & 226 & 302 & 70 \\
		\noise{} $\gg$ \baseline{}  & 178 & 232 & 97 \\				
		\noise{} $\gg$ \google{}     & 218 & 315 & 65 \\
		\baseline{} $\gg$ \google{}     & 173 & 302 & 123 \\
	\end{tabular}
	\vspace{-.15cm}
	\caption{In-house human evaluation (``$\gg$'' means better with $p\leq{}0.05$).
	The 4 models \emph{\baseline{}}, \emph{\google{}}, \emph{\tags{}} and \emph{\noise{}} correspond respectively to rows 2 (UGC with inline case), 3 (Google Translate), 6 (Combination of BT, PE and tags) and 8 (Same as 6 with natural noise) in Table~\ref{table:BLEU_combination}.}
	\label{table_NLE_results}
\end{table}

\section{Conclusion}
\label{section:conclusion}

\vspace{-.05cm}

We presented a new parallel corpus of user reviews of restaurants, which we think will be valuable to the community.
We proposed combinations of multiple techniques for robustness and domain adaptation, which address particular challenges of this new task.
We also performed an extensive evaluation to measure the improvements brought by these techniques.

According to BLEU, the best single technique for domain adaptation is fine-tuning.
Corpus tags also achieve good results, without degrading performance on a general domain.
Back-translation helps, but only with sampling or tags.
The robustness techniques (natural noise, factored case, rare character placeholder) do not improve BLEU.

While our models are promising, they still show serious errors when applied to user-generated content: missing negations, hallucinations, unrecognized named entities, insensitivity to context.\footnote{See additional examples in Appendix.} This suggests that this task is far from solved.

We hope that this corpus, our natural noise dictionary, model outputs and human rankings will help better understand and address these problems.
We also plan to investigate these problems on lower resource languages, where we expect the task to be even harder.

\let\OLDthebibliography\thebibliography
\renewcommand\thebibliography[1]{
	\OLDthebibliography{#1}
	\setlength{\parskip}{3pt}
	\setlength{\itemsep}{2pt}
}

\bibliography{paper}
\bibliographystyle{acl_natbib}

\clearpage
\section*{Appendix}

\begin{table}[H]
	\centering
	\begin{tabular}{lccc}
		Corpus & Emojis & Capitalized & Typos \\
		\hline
		\foursq{}-test & 0.17 & 0.14 & 3.3 \\
		MTNT-test & 0.02 & 0.18 & 3.8 \\
	\end{tabular}
	\caption{Noise comparison between \foursq{}-test and MTNT-test \cite{michel_mtnt:_2018}.
		Emojis, all-uppercase words (not counting acronyms) and spelling + grammar mistakes (according to MS Word) per 100 tokens.}
	\label{table:MTNT_comparison}
\end{table}

\begin{table}[H]
	\centering
	\begin{tabular}{l|c|c}
		\multirow{2}{*}{Model} & Foursq. & MTNT- \\
		& valid & test \\
		\hline
		\multicolumn{3}{c}{\citet{berard:2019:WMT}} \\
		\hline
		WMT (Inline case) & -- & 39.1 \\
		+ MTNT domain adaptation & -- & 44.3 \\
		+ Ensemble & -- & \textbf{45.7} \\
		\hline
		\multicolumn{3}{c}{Our models (single)} \\
		\hline
		WMT (Cased) & 24.3 & 39.0 \\ 
		UGC (Cased) & 30.4 & 41.5 \\ 
		UGC (Inline case) & 29.3 & 41.6 \\ 
		UGC $\oplus$ BT + FT & 33.7 & 44.5 \\ 
		UGC $\oplus$ BT $\oplus$ PE + tags & 33.7 & \textbf{44.9} \\ 
		Nat noise $\oplus$ BT + FT & \textbf{33.8} & 44.6 \\ 
		Nat noise $\oplus$ BT $\oplus$ PE + tags & 33.4 & \textbf{44.9} \\ 
	\end{tabular}
	\caption{Comparison of our models against the winner of the WMT 2019 Robustness Task on the MTNT test set (similar robustness challenges but different domain). We also give cased BLEU of our models on \foursq{}-valid. Results on \foursq{}-test are shown in the paper.}
\end{table}

\paragraph{Large-Scale monolingual evaluation}
We conducted a larger scale monolingual evaluation using Amazon Mechanical Turk (AMT), as reported in Table~\ref{table_AMT_results}.
We evaluated the translations of 1800 test sentences.
To filter poor quality work, which occurs frequently in our experience, we also created gold questions by selecting 40 additional sentences for which we built 3 fake translations each, whose ranking was intentionally unambiguous and easy.
We created HITs (Human Intelligence Tasks) of 10 sentences each, of which 3 sentences were gold questions.
Workers were also required to have at least 98\% task approval rate on AMT and 1000 tasks approved.
We aimed for 6 submissions per HIT from 6 different workers.
Compared to the in-house evaluation, the inter-judge agreement was low (Kappa of 0.15).

\begin{table}[H]
	\centering
	\begin{tabular}{l|c|c|c}
		Pairs & Win & Tie & Loss \\
		\hline
		\noise{} $\gg$ \tags{}         & 1939 & 7414 & 1667 \\
		\noise{} $\gg$ Base            & 2718 & 6108 & 2178 \\
		\noise{} $\gg$ \google{}       & 3008 & 5801 & 2173 \\
		\tags{} $\gg$ \baseline{}      & 2657 & 6110 & 2225 \\				
		\tags{} $\gg$ \google{}        & 2950 & 5794 & 2234 \\
		\baseline{}  $\gg$ \google{}   & 2205 & 6918 & 1889 \\
	\end{tabular}
	\caption{Large-scale Human Evaluation on Amazon Mechanical Turk (``$\gg$'' means $p\leq{}0.01$).
		The 4 models \emph{\baseline{}}, \emph{\google{}}, \emph{\tags{}} and \emph{\noise{}} correspond respectively to rows 2 (UGC with inline case), 3 (Google Translate), 6 (Combination of BT, PE and tags) and 8 (Same as 6 with natural noise) in Table~\ref{table:BLEU_combination}.}
	\label{table_AMT_results}
\end{table}

Both human evaluations agree and are consistent with the BLEU evaluation, except for the impact of natural noise, where the AMT evaluation found a significant improvement.

\begin{table}[H]
	\begin{tabular}{l|c|c|c|c}
		Evaluation & \# Tasks & \# Ties & \% Ties & Kappa  \\
		\hline
		In-house   & 12       &  3588   & 57\% & 0.47 \\
		AMT	       & 1321 & 65988 & 58\% & 0.15 \\
	\end{tabular}
	\caption{Size of the human evaluations.
		AMT: Amazon Mechanical Turk.
		The AMT kappa (inter-judge agreement) is very low, while the in-house kappa is moderate.}
	\label{Table_human_eval_figures}
\end{table}

\clearpage

\newcolumntype{P}{>{\ttfamily\small}p}
\begin{table*}
	\centering
	\begin{tabular}{lP{13cm}}
		SRC & On s'y sent comme \textbf{\textcolor{red}{a}} la maison ! <s> Équipe de serveurs très \verb|sympa|! <s> \textbf{\textcolor{red}{Goutez}} au burger \textbf{\textcolor{red}{LE Retour d'Hervé}}, il est \textbf{\textcolor{red}{a tomber :-)}}\\
		REF & It feels like home!! <s> Team of waiters very nice! <s> Taste the burger LE Retour d'Hervé, it's to die for :-) \\
		Type & Bar, Bistro \\
		Location & Paris, FR \\
		Rating & 8.29 \\
		\hline
		SRC & Je conseille le crumble fraise/rhubarbe \textbf{\textcolor{red}{CHAUD}}. <s> C'est délicieux !!\\
		REF & I recommend the strawberry/rhubard crumble HOT. <s> It's delicious!!\\
		Type & Bakery, Breakfast Spot \\
		Location & Brussels, BE \\
		Rating & 8.88 \\
		\hline
		SRC & Très bons burgers, cheesecake \textbf{\textcolor{red}{à tomber par terre....}} <s> Sans oublier <NAME>, <NAME> et <NAME> en un mot \textbf{\textcolor{red}{CHAR-MANTS}}!\\
		REF & Very good burgers, cheesecake to die for... <s> Not to mention <NAME>, <NAME> and <NAME>: in a word CHAR-MING!\\
		Type & American Restaurant \\
		Location & Paris, FR \\
		Rating & - \\
		\hline
		SRC & Friterie sympathique collée \textbf{\textcolor{red}{au Grand}} Boulevards. <s> On retrouve les incontournables frites belges. <s> \textbf{\textcolor{red}{Elle}} sont \textbf{\textcolor{red}{DELICIEUSESEMENT}} grosses comme on \textbf{\textcolor{red}{aiment :) a}} tester. <s> Ouverture tardive le \textbf{\textcolor{red}{we}}.\\
		REF & Friendly chip shop stuck to Grand Boulevards. <s> We find the essential Belgian fries. <s> They are DELICIOUSLY big as we like them :) to test. <s> Late opening on the weekend.\\
		Type & Belgian Restaurant, Fast Food Restaurant \\
		Location & Paris, FR \\
		Rating & 7.91 \\
		\hline
		SRC & Que de \textbf{\textcolor{red}{bon souvenir , fillet}} de boeuf \textbf{\textcolor{red}{au patte}}. <s> Merci pour \textbf{\textcolor{red}{l accueille Mr}} <NAME> \\
		REF & Great memories, beef fillet with pasta. <s> Thank you for being so welcoming Mr <NAME>\\
		Type & Café, Pizza Place \\
		Location & Libreville, GA \\
		Rating & 8.21 \\
		\hline
		SRC & La \textbf{\textcolor{red}{carte}} est souvent enrichie. <s> La gérance est \textbf{\textcolor{red}{top}}.\\
		REF & The menu is often supplemented. <s> The management is top notch.\\
		Type & Sushi Restaurant \\
		Location & Sid'Bou Said, TN \\
		Rating & 7.70 \\
	\end{tabular}
	\label{appendix:4SQ_examples}
	\caption{Examples of challenging examples from \foursq{}-PE. We show the full reviews with sentence delimiters (\texttt{<s>}) and metadata. The words that contain typos or that could cause trouble to a regular NMT model are shown in bold red.}
\end{table*}

\label{appendix:mt_outputs}
\begin{table*}
	\centering
	\begin{tabular}{lP{12cm}}
		SRC & Le meilleur resto de Belleville, DE LOIN! \\
		REF & The best restaurant in Belleville, BY FAR! \\
		Cased & Best restaurant in Belleville, DE LOIN! \\
		Inline case & The best restaurant in Belleville, BY FAR! \\
		\hline
		SRC & ESCALOPE DE VEAU MONTAGNARDE à tomber, et à ne plus pouvoir se lever de sa chaise \\
		REF & ESCALOPE DE VEAU MONTAGNARDE is an absolute knock out and you'll have difficulty recovering \\
		Cased & Falling down and not being able to get up from his chair \\
		Inline case & ESCALOPE OF MOUNTAIN CALF to fall, and not be able to rise from his chair \\
	\end{tabular}
	\caption{Examples of sentences from \foursq{}-test with capitalized words, where default (cased) MT gets the translation wrong and inline case helps.}
\end{table*}

\begin{table*}
	\centering
	\begin{tabular}{lP{12cm}}
		SRC & \textbf{\textcolor{red}{Bcp}} de choix, peut-être Trop :-) \\
		REF & \textbf{\textcolor{red}{Plenty}} of choice, maybe too much of it :-) \\
		Inline case & \textbf{\textcolor{red}{Bcp}} of choice, maybe Too much :-) \\
		Natural noise & \textbf{\textcolor{red}{A lot}} of choices, maybe Too much :-) \\
		\hline
		SRC & Service \textbf{\textcolor{red}{loooooonnnng}}. \\
		REF & \textbf{\textcolor{red}{Looooooong}} wait. \\
		Inline case & Service \textbf{\textcolor{red}{loooooonnnng}}. \\
		Natural noise & \textbf{\textcolor{red}{Long}} service.
	\end{tabular}
	\caption{Examples of sentences from \foursq{}-test with noisy spelling (in bold red), where training with source-side natural noise helps.}
\end{table*}

\begin{table*}
	\centering
	\begin{tabular}{lP{12cm}}
		SRC & \textbf{\textcolor{red}{Carte}} attractive et pas excessive. \\
		REF & Nice \textbf{\textcolor{red}{menu}} and not over the top. \\
		Inline case & Attractive and not excessive \textbf{\textcolor{red}{card}}. \\
		BT + FT & Attractive \textbf{\textcolor{red}{menu}} and not excessive. \\
		\hline
		SRC & \textbf{\textcolor{red}{Cuisine}} pas originale, service passable, mais l'endroit est joli ! \\
		REF & Not very original \textbf{\textcolor{red}{food}}, acceptable service, but the place itself is beautiful! \\
		Inline case & Not an original \textbf{\textcolor{red}{kitchen}}, fair service, but the place is nice! \\
		BT + FT & \textbf{\textcolor{red}{Food}} not original, service passable, but the place is nice!
	\end{tabular}
	\caption{Examples of sentences from \foursq{}-test with polysemous words (in bold red), where domain adaptation helps (with \foursq{}-PE fine-tuning and back-translation).}
\end{table*}

\begin{table*}
	\centering
	\begin{tabular}{lP{10cm}|p{5cm}}
		SRC & Les frittes \textbf{\textcolor{red}{boff}} mais leurs burger, une tuerie! & \multirow{3}{5cm}{Typo and slang (``bof'')} \\
		REF & The fries are \textbf{\textcolor{red}{meh}}, but the burgers, to die for! & \\
		MT & The fries are \textbf{\textcolor{red}{great}} but their burgers are to die for!& \\
		\hline
		SRC & Le \textbf{\textcolor{red}{merveilleux}} du \textbf{\textcolor{red}{Merveilleux}} c'est merveilleux... & \multirow{3}{5cm}{``merveilleux'' is a pastry, ``Merveilleux'' is a pastry shop (named entity).} \\
		REF & The \textbf{\textcolor{red}{merveilleux}} at \textbf{\textcolor{red}{Merveilleux}} is marvelous... & \\
		MT & The \textbf{\textcolor{red}{wonderful}} of the \textbf{\textcolor{red}{Wonderful}} it's wonderful... & \\
		\hline
		SRC & La \textbf{\textcolor{red}{souris d'agneau}} est délicieuse ! & \multirow{3}{5cm}{Dish name (translated literally)} \\
		REF & The \textbf{\textcolor{red}{lamb shank}} is delicious! & \\
		MT & The \textbf{\textcolor{red}{lamb mouse}} is delicious! & \\
		\hline
		SRC & La quantité 5 raviolis \textbf{\textcolor{red}{qui se battent}} pour 12.70 euros. & \multirow{3}{5cm}{Idiomatic expression (``qui se battent en duel'')} \\
		REF & Poor quantity, 5 raviolis or so for 12.70 Euros. & \\
		MT & The quantity 5 dumplings \textbf{\textcolor{red}{that fight}} for 12.70 euros. & \\
		\hline
		SRC & Après le \textbf{\textcolor{red}{palais du facteur}} nous voici à \textbf{\textcolor{red}{la halte}} qui est un restaurant correct. & \multirow{3}{5cm}{Named entities (``Palais Idéal du Facteur Cheval'' and ``La Halte du Facteur'')} \\
		REF & After the \textbf{\textcolor{red}{Palais du Facteur}} we stopped at \textbf{\textcolor{red}{La Halte}}, which is a reasonable restaurant. & \\
		MT & After the \textbf{\textcolor{red}{mailman's palace}} here we are at the \textbf{\textcolor{red}{rest stop}} which is a decent restaurant. & \\
	\end{tabular}
	\caption{Examples of bad translations by our best model (Noise $\oplus$ BT $\oplus$ PE + tags). All examples are from \foursq{}-test, except for the last one, which is from SemEval.}
\end{table*}

\end{document}